# A Multimodal Fusion Network for Student Emotion Recognition Based on Transformer and Tensor Product

Ao Xiang
School of Computer Science & Engineering
University of Electronic Science and Technology of China
Chengdu, Sichuan, China

Zongqing Qi*
Computer Science
Stevens Institute of Technology
Hoboken NJ, U.S
*zongqingqi@gmail.com

Han Wang
Financial Mathematics
University of Southern California
LA, U.S

Qin Yang
School of Integrated Circuit Science and Engineering
University of Electronic Science and Technology of China
Chengdu, Sichuan, China

Danqing Ma
Computer Science
Stevens Institute of Technology
Hoboken NJ, U.S

*Abstract*—This paper introduces a new multi-modal model based on the Transformer architecture and tensor product fusion strategy, combining BERT's text vectors and ViT's image vectors to classify students' psychological conditions, with an accuracy of 93.65%. The purpose of the study is to accurately analyze the mental health status of students from various data sources. This paper discusses modal fusion methods, including early, late and intermediate fusion, to overcome the challenges of integrating multi-modal information. Ablation studies compare the performance of different models and fusion techniques, showing that the proposed model outperforms existing methods such as CLIP and ViLBERT in terms of accuracy and inference speed. Conclusions indicate that while this model has significant advantages in emotion recognition, its potential to incorporate other data modalities provides areas for future research.

## I. INTRODUCTION

The impact of students' psychological states on their overall well-being, academic performance, and physical and mental development is a critical concern in the field of education. Assessing the psychological status of students has traditionally been conducted through interviews with psychologists. However, this method is noted for its inefficiency. The digital age has brought about a significant transformation in educational systems worldwide. This is characterized by the explosive growth of educational informatization and the accumulation of vast amounts of educational big data. As a result, educational data mining (EDM) has become a concept that is gaining strength. Educational Data Mining (EDM) is a specialized branch of data mining that is focused on applying data mining techniques and models to uncover hidden patterns in educational data sets. Integrating data mining technology into educational systems has significantly enhanced the capability to monitor and identify students' mental health status. By using EDM, educators and psychologists can more efficiently analyze educational data to identify at-risk students and provide timely support for optimizing students' living conditions and learning efficiency, ultimately fostering their holistic development. However, the data on student mental health primarily comes from questionnaire surveys, which may not be sufficient to identify early warning signs of abnormal psychological conditions due to issues such as data sparsity, accuracy, and timeliness.

Concurrently, the ascendancy of the internet has catalyzed the widespread application of artificial intelligence (AI) technologies across various sectors, including finance [2], medicine [3], and water conservancy construction [4]. Educators and AI researchers are now leveraging machine learning technologies to extract information from diverse data modalities, such as images and texts. By analyzing student thoughts and perspectives from various data sources, it is possible to accurately identify and understand their mental health states. Educators and AI researchers are now leveraging machine learning technologies to extract information from multiple data modalities, such as images and text. By analyzing student thinking and perspectives from different data modalities, it is feasible to more accurately identify and understand student mental health and emotional states.

Initially, traditional machine learning methods were primarily used for analyzing student emotions in multimodal approaches. For instance, researchers like Cook et al. [5] have employed text analysis to evaluate students' feedback on teaching, creating term-document matrices and using algorithms such as random forests for teacher evaluation purposes. However, the effectiveness of these traditional machine learning approaches remained suboptimal. With the emergence of deep learning, there has been a significant improvement in the predictive performance of multimodal fusion models based on neural networks. For example, Wang et al.[6] used the xlnet large model to extract and establish a deep neural network classification model for classification. The effect has been significantly improved compared to the traditional machine learning model. However, there are still many challenges in extracting features relevant to students' mental health and emotions, and in integrating multimodal information from a variety of student characteristics.

This study builds a multi-modal model based on Transformer architecture and tensor product fusion strategy to identify students' psychological and emotional conditions. This model also classifies students' psychological conditions to achieve behavioral early warning of students' abnormal psychological conditions. To extract text vectors, the model employs the BERT model, and for image vectors, it uses ViT. After fusing at the tensor level, the data is passed to the MLP neural network, resulting in an accuracy of 93.65%.

## II. RELATED WORK

Modal fusion methods typically include early fusion and late fusion. Early fusion, also known as feature fusion, refers to the merging of data or features from multiple modalities prior to model input, allowing the model to process the fused data from the outset so that it can directly learn and process the interactions between the different modalities. Late fusion, also known as decision-level fusion, refers to the merging of independent judgments or outputs from different modalities during the model's final decision stage [7]. The advantage of early fusion is its ability to merge the data at the beginning, which makes subsequent operations easier, but at the cost of forcing the compression and suppression of different feature scales and modality distribution differences at the beginning of the fusion phase, which may result in incomplete use of the information. On the contrary, late fusion allows each modality to be trained individually applying the most appropriate processing and learning methods, which can take good care of different types of features, but after training different models individually, only integrating the outputs in the final stage may be difficult to capture the characteristics of deep interactions between different modalities, resulting in less-than-ideal generalization performance. Therefore, choosing a suitable fusion method to detect student depression remains a challenge.

In the face of this challenge, intermediate fusion emerged. Intermediate's fusion approach involves merging features from different modalities at some intermediate layer of the model. This fusion approach tries to balancing the features of early and late fusion by performing a portion of the feature merging operation first at some stage within the model, which enables the fusion operation to be placed in multiple layers of the deep learning model, thus dealing with the problem of information interaction in a more flexible way. The two most commonly used approaches are simple splicing and approaches based on attention mechanisms[8]. The simple operation fusion approach straightforwardly integrates feature vectors from different modalities can be achieved by simple operations such as splicing and weighted summation. Let the subsequent network layer will automatically adapt itself to such operations. The attention mechanism approach, on the other hand, is a set of scalar weight vectors dynamically generated by the model at each time step, allowing different modal vectors to be weighted and summed [9] . The multiple output heads of this set of attention can be dynamically generated with the weights to be used in the summation, so that ultimately additional weight information is preserved at the time of splicing.

## III. METHODOLOGY

### A. Dataset

We collected social network software messaging transcripts from WeChat in a university in Guangdong, China, including text and emoticon images, and we aligned and matched the text of the next sentence immediately after the emoticon image as the timestamp of the image. The data was manually annotated by an emotion recognition expert according to the Eysenck Personality Questionnaire[11], which categorized the data as positive, negative, or neutral. Finally, we have collected a total of 262 matching image text messages from 191 users.

### B. Modal Processing

Because our data collection is relatively small, training from scratch may be a great challenge to the model's generalization ability, therefore, we adopt the BERT model to vectorize the text, and CLS tags are used as the text-sentence sentiment output representation. The BERT (Bidirectional Encoder Representation from Transformers) model is a pre-trained model proposed by Google in 2018 [12], which utilizes a network on the encoder side of the Transformer that is able to collect information about the word's relevant context and encode it in rich vectors representing that context, as shown in Figure 1. The encoding vectors of the inputs to the BERT are the units of five embedded features. The input vector of BERT is composed of five embedding features, which are token embedding, position embedding, and segment embedding. In the token embedding, the first element is CLS, and the last layer of the BERT model corresponds to the vector as the semantic representation of the whole sentence. This token with no obvious semantic information will fuse the semantic information of each word in the text more fairly, to better represent the semantics of the whole sentence.

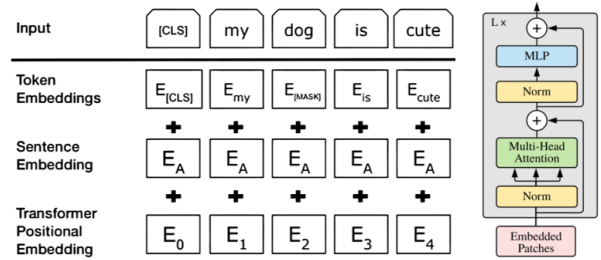

Fig. 1 Structure of BERT embedding

For the image information, we capture it using the ViT model, which is a pre-trained model also proposed by Google [13]. It is modeled after operations in natural language processing and aims to convert image data into sequences. Its mechanism of operation is shown in Figure 2. Similar to BERT, it is designed with position embedding and CLS embedding, and the class token in the output of the transformer encoder is treated as the model's encoded features of the input image.

After extracting the correlation vectors, we use the tensor product method using tensor fusion network [14] to fuse them. Tensor fusion network is a neural network architecture for multimodal data processing. It contains three dimensions and is an end-to-end network structure. The core concept is to fuse the feature representations of various modalities by tensor product to form a joint feature representation that be able to contain all possible interactions between modalities as represented in Figure 3. The final structure of this study is shown in Figure 4.

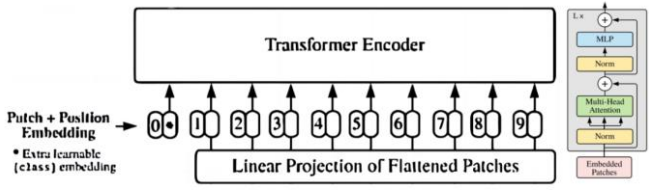

Fig. 2 Structure of ViT embedding

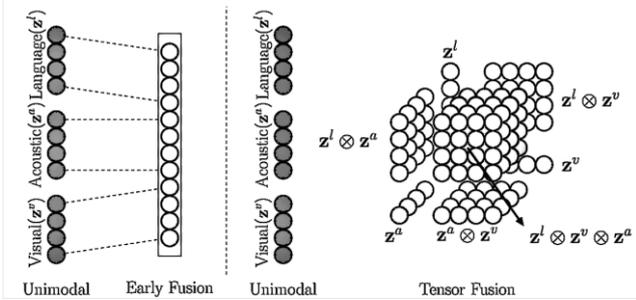

Fig .3 Structure of tensor fusion network

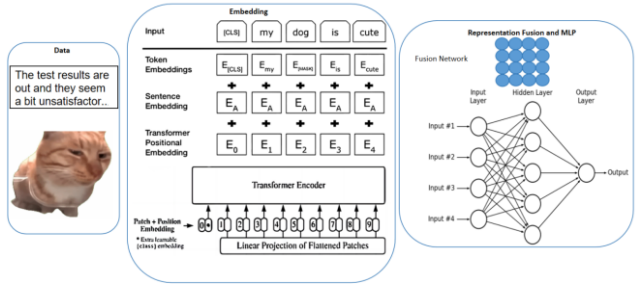

Fig. 4 The fusion network based on transformer and tensor product

*C. Evaluation Metrics*

In the classification task, two commonly used evaluation metrics, Accuracy and AUC (Area Under the ROC Curve), are applied in this paper. Accuracy measures the proportion of all predictions that are correctly predicted, that is, the proportion of the total number of samples that are correctly predicted by the model as positive and negative categories to the total number of predicted samples. AUC is under various categorization thresholds and it is the area under the ROC curve, which is a curve obtained by plotting the True Positive Rate (TPR) versus the False Positive Rate (FPR) at varying thresholds, where the True Positive Rate refers to the proportion of samples correctly categorized by the model as a positive class among all actual positive classes; and the False Positive Rate is the proportion of all actual negative samples that are misclassified as positive by the model.

## IV. EXPERIMENT

We will observe our results through two types of experiments. The first is an ablation experiment, where we will split the various basic models of our model. The second type is a comparison between mainstream models.

Table I presents the performance of different base models and fusion techniques in recognizing emotions in online textual communication between students, specifically by conducting two sets of ablation studies. The first set is aimed at evaluating the ability of models in different modalities in emotion recognition, while the second set evaluates the performance after applying various fusion methods in a multimodal framework. The results of the initial ablation study indicate that unimodal approaches have an accuracy rate of over 90% in discerning emotions, indicating a competent ability to accurately classify the emotional content of students' conversations over chat applications. However, the study also shows that ViT's performance is 1.76% lower than BERT's accuracy, suggesting that ViT may be playing a less effective role in this specific context of emotion classification. In addition to the technical reasons, this may be due to the fact that the student population may have certain preferences for emoticons in instant messaging software, which are different from the pre-training data of Vit, which makes the more abstract expressions may have a lower success rate for Vit to recognize in comparison to the language.

Further, by straightforwardly concatenating the outputs from BERT and ViT, the fused model exhibits an enhancement, reaching an accuracy of 92.81% and an AUC of 91.60%.This improvement highlights the advantages of a more direct and simple fusion, which allows for more extensive interaction between the two modalities, thus improving the model's performance. In addition, employing a dot product approach for fusion yields a model with an accuracy of 92.29% and an AUC of 91.36%, outperforming the uni-modal frameworks but slightly trailing behind the concatenation fusion. This method of dot product is based on cosine similarity. This outcome may indicate the dot product's tendency to overemphasize similarities, which can result in a loss of nuanced information across modalities, especially when integrating detailed and complex interactions. This hypothesis is corroborated by the results achieved through tensor product fusion, which constructs a higher-dimensional feature space during the fusion process and accounts for all possible interactions among features. As a result, the model's efficiency is significantly improved, achieving an accuracy of 93.65% and an AUC of 96.58%.

TABLE I ABLATION EXPERIMENT RESULTS

| Model | Fusion Type | Accuracy | Precision | AUC |
|---|---|---|---|---|
| BERT | None | 0.9190 | 0.9018 | 0.9064 |
| ViT | None | 0.9014 | 0.8946 | 0.8995 |
| BERT-ViT | Concat | 0.9281 | 0.9205 | 0.9160 |
| BERT-ViT | Dot product | 0.9229 | 0.9201 | 0.9136 |
| BERT-ViT (Ous) | Tensor product | 0.9365 | 0.9367 | 0.9658 |

Table II presents the results of the multi-model performance. Our model is compared with other mainstream models, such as the widely used CLIP and ViLBERT model and measured the inference speed of different models. The CPU of the device is Intel i3-8100, the memory is 64gb, and the batch is 128. Empirical evidence has shown that CLIP's performance is not as good as our model for several reasons. Firstly, the CLIP model employs cosine similarity for fusion, but ablation studies already proved that this method of fusion is not as effective as the vector product. The disadvantage of CLIP is that if the data used for inference is really far from the training data, then the generalization ability of CLIP will become poor. The images sent in student instant messaging are often abstract artistic recreations, which are unlikely to be included in CLIP's pre-training data set. This could be a contributing factor to CLIP's performance falling short compared to our model. Another comparative model, ViLBERT (Vision-and-Language BERT), is an end-to-end multimodal model based on improved BERT. ViLBERT is slightly 0.24% higher in accuracy than our BERT-ViT, but the inference speed is higher than compared to our model, it is much longer. Although the nature of ViLBERT's end-to-end architecture can reduce the inference time to some extent, cross-attention and complex multi-head mechanisms cause its inference speed to be unsatisfactory. As a result, our model is more suitable for the downstream tasks of emotion recognition.

TABLE II Model comparison results

| Model | Fusion Type | Accuracy | Precision | Inference speed |
|---|---|---|---|---|
| BERT-ViT(Ous) | Tensor product | 0.9365 | 0.9367 | 12.81s |
| CLIP | Cosine similarity | 0.9295 | 0.9231 | 11.18s |
| ViLBERT | Cross attention | 0.9389 | 0.9366 | 13.79s |

V. Conclusion

This paper introduces a multimodal emotion recognition network based on pre-trained models such as BERT and Vit for representation extraction, and uses tensor product for multimodal fusion. The model performs well on student datasets, and ablation experiments confirm its superiority. The pre-trained large model with the same transformer structure makes the dimensions of the semantic feature output consistent and can achieve great results with less data. The fusion form of tensor product enables deep interaction between different modalities and captures more subtle information. However, it is not an end-to-end model and employs two pre-training models, which means that the network is obviously resource-intensive. This study only discusses and analyzes two types of modes. If additional modes are included, such as student quantified features or audio modalities, performance in Modalities can be expected to be further improved, In future research, further improvements can be made towards lightweight and high performance.